\documentclass[twoside, 10pt]{article}
\usepackage{times,amsmath,amstext,amssymb}
\usepackage{graphicx}
\usepackage{times}
\usepackage[hang,scriptsize]{subfigure}
\usepackage{amsmath}  
\usepackage{amssymb}  
\usepackage{graphics} %
\usepackage{latexsym}
\usepackage{amsfonts}
\usepackage{stmaryrd}
\usepackage{fancyhdr}

\fancypagestyle{plain}
 {%
     \fancyhf{} %
     \fancyfoot[L]{ \textit {INFORMATION \& SECURITY. An International Journal,
      Vol. 20, 2006, 1-22}}   

 }
 
\usepackage[T1]{fontenc}
\usepackage{epsfig}
\usepackage{amsmath}
\usepackage{rotating}

\newcommand{\bel}{\mathrm{bel}}
\newcommand{\pl}{\mathrm{pl}}
\newcommand{\betP}{\mathrm{betP}}
\newcommand{\Bel}{\mathrm{Bel}}
\newcommand{\Pl}{\mathrm{Pl}}
\newcommand{\GPT}{\mathrm{GPT}}

\input{epsf}
\begin{document}
\thispagestyle{plain}
\input{epsf}         
\date{}


\begin{center}  

 {\Large { \textbf { HUMAN EXPERTS FUSION FOR IMAGE CLASSIFICATION}}}  

 \vspace{1cm}

 { \large  { Arnaud MARTIN and Christophe OSSWALD}}   

\end{center}

\baselineskip 13pt




\begin{abstract}
In image classification, merging the opinion of several human experts is very important for different tasks such as the evaluation or the training. Indeed, the ground truth is rarely known before the scene imaging. We propose here different models in order to fuse the informations given by two or more experts. The considered unit for the classification, a small tile of the image, can contain one or more kind of the considered classes given by the experts. A second problem that we have to take into account, is the amount of certainty of the expert has for each pixel of the tile. In order to solve these problems we define five models in the context of the Dempster-Shafer Theory and in the context of the Dezert-Smarandache Theory and we study the possible decisions with these models.

\end{abstract}

\section*{Introduction}

Fusing the opinion of several human experts, also known as the experts fusion problem, is an important question in the image classification field and very few studied. Indeed, the ground truth is rarely known before the scene has been imaged; consequently, some experts have to provide their perception of the images in order to train the classifiers (for supervised classifiers), and also to evaluate the image classification. In most of the real applications, the experts cannot provide the different classes on the images with certitude. Moreover, the difference of experts perceptions can be very large, and so many parts of the images have conflicting information. Thereby, only one expert {\it reality} is not reliable enough, and experts fusion is required.

Image classification is generally done on a local part of the image (pixel, or most of the time on small tiles of {\it e.g.} 16$\times$16 or 32$\times$32 pixels). Classification methods can usually be described into three steps. First, significant features are extracted from these tiles. Generally, a second step in necessary in order to reduce these features, because they are too numerous. In the third step, these features are given to classification algorithms. The particularity in considering small tiles in image classification is that sometimes, more than one class can co-exist on a tile. 

\begin{figure}[htb]
\includegraphics[height=4.7cm]{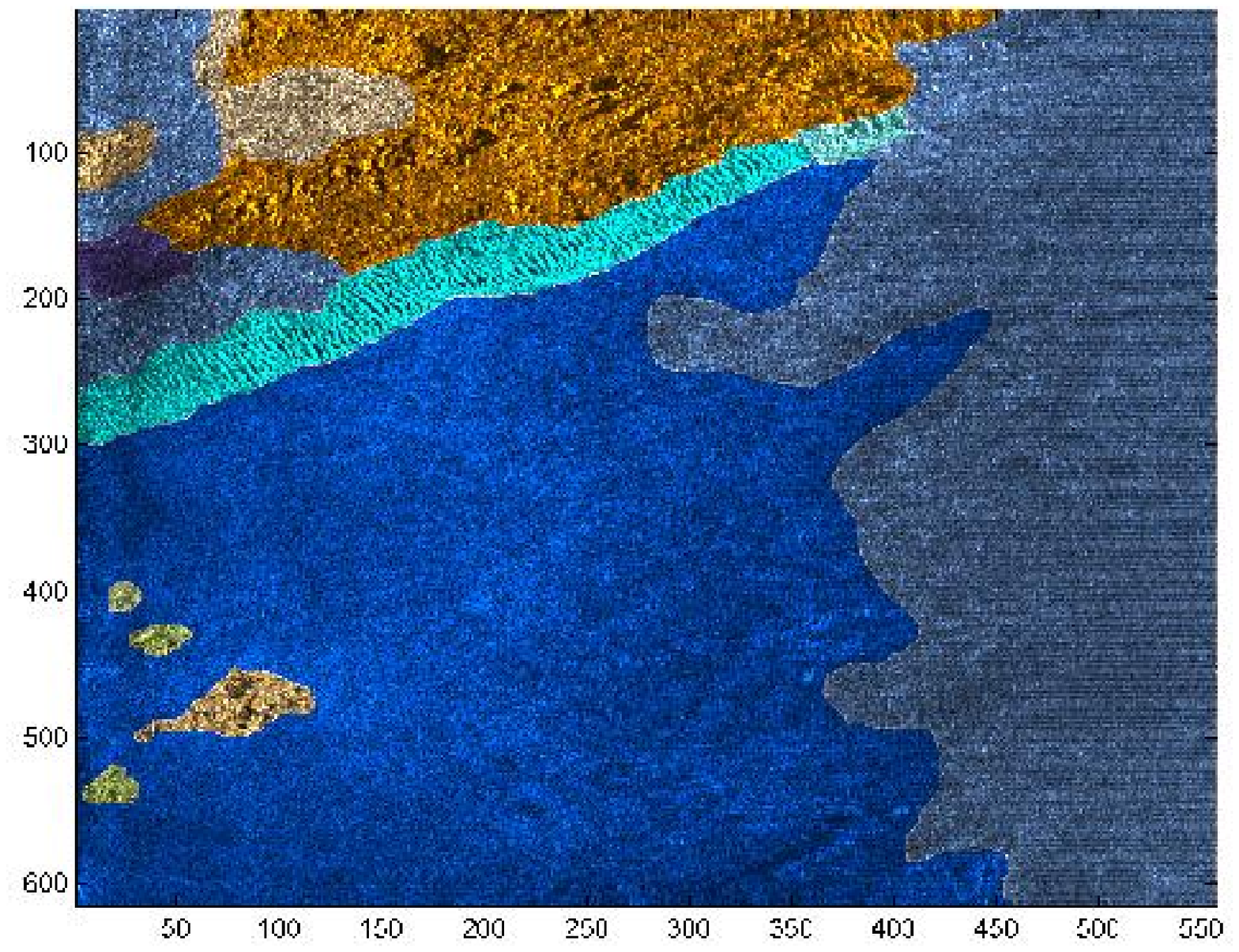}
\includegraphics[height=4.7cm]{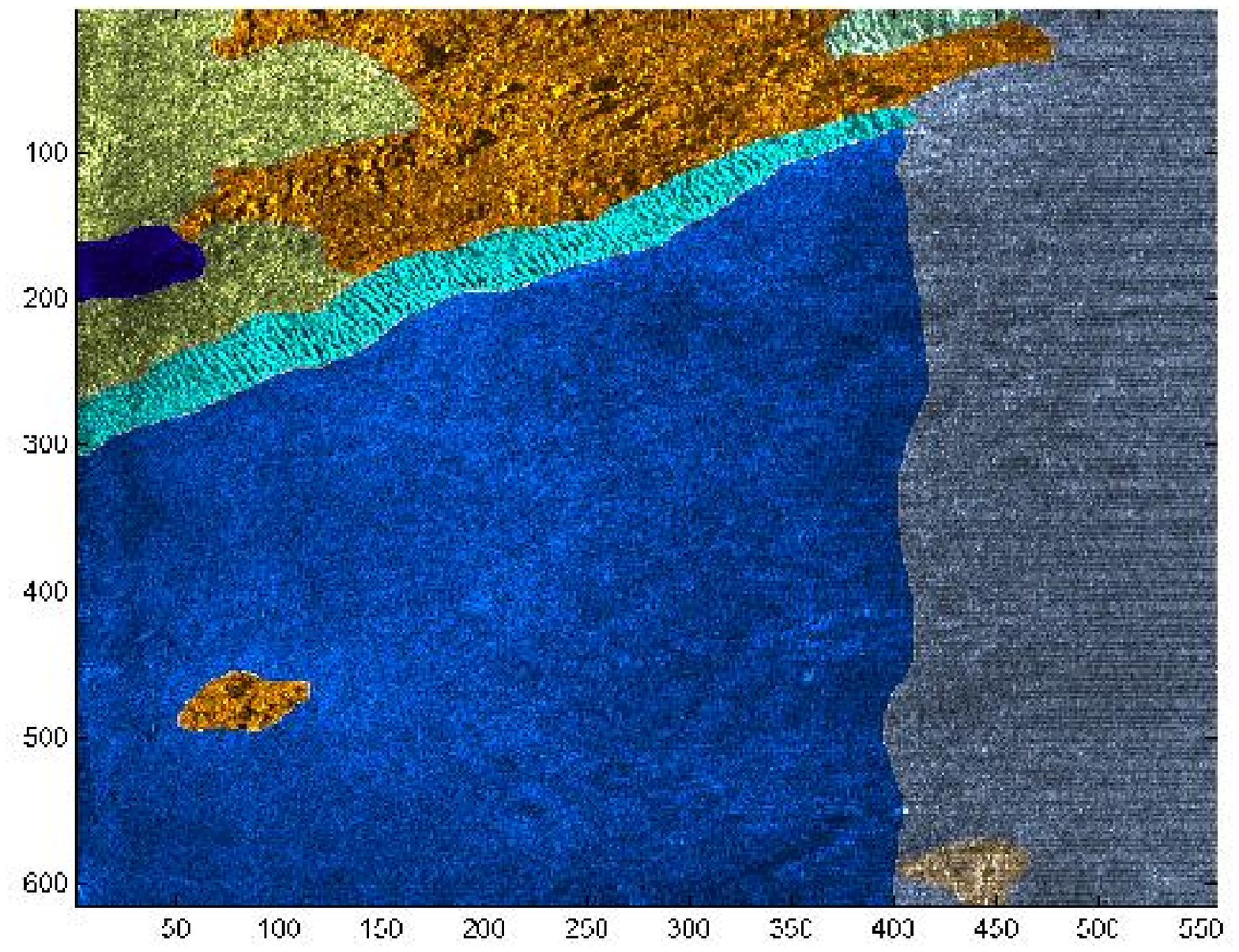}
\vspace{-0.5cm}
\caption{Segmentation given by two experts.}
\label{expert}
\end{figure}

An example of such an image classification process is seabed characterization. This serves many useful purposes, {\it e.g} help the navigation of Autonomous Underwater Vehicles or provide data to sedimentologists. In such sonar applications, which serve as examples throughout the paper, seabed images are obtained with many imperfections \cite{Martin05}. Indeed, in order to build images, a huge number of physical data (geometry of the device, coordinates of the ship, movements of the sonar, etc.) are taken into account, but these data are polluted with a large amount of noises caused by instrumentations. In addition, there are some interferences due to the signal traveling on multiple paths (reflection on the bottom or surface), due to speckle, and due to fauna and flora. Therefore, sonar images have a lot of imperfections such as imprecision and uncertainty; thus sediment classification on sonar images is a difficult problem. In this kind of applications, the reality is unknown and different experts can propose different classifications of the image. Figure \ref{expert} exhibits the differences between the interpretation and the certainty of two sonar experts trying to differentiate the type of sediment (rock, cobbles, sand, ripple, silt) or shadow when the information is invisible. Each color corresponds to a kind of sediment and the associated certainty of the expert for this sediment expressed in term of sure, moderately sure and not sure. Thus, in order to learn an automatic classification algorithm, we must take into account this difference and the uncertainty of each expert. For example,  how a tile of rock labeled as {\it not sure} must be taken into account in the learning step of the classifier and how to take into account this tile if another expert says that it is sand? Another problem is: how to take into account the tiles with more than one sediment?

Many fusion theories can be used for the experts fusion in image classification such as voting rules \cite{Xu92,Lam97}, possibility theory \cite{Zadeh78,Dubois88a}, belief function theory \cite{Dempster67,Shafer76}. In our case, experts can express their certitude on their perception. As a result, probabilities theories such as the Bayesian theory or the belief function theory are more adapted. Indeed, the possibility theory is more adapted to imitate the imprecise data whereas probability-based theories is more adapted to imitate the uncertain data. Of course both possibility and probability-based theories can imitate imprecise and uncertain data at the same time, but not so easily. That is why, our choice is conducted on the belief function theory, also called the Dempster-Shafer theory (DST) \cite{Dempster67, Shafer76}. We can divide the fusion approach into four steps: the belief function model, the parameters estimation depending on the model (not always necessary), the combination, and the decision. The most difficult step is presumably the first one: the belief function model from which the other steps follow. 

Moreover, in real applications of image classification, experts conflict can be very large, and we have to take into account the heterogeneity of the tiles (more than one class can be present on  the tile). Consequently, the Dezert-Smarandache Theory (DSmT) \cite{Smarandache04}, an extension of the belief function theory, can fit better to our problem of image classification if there is conflict. Indeed, considering the space of discernment $\Theta=\{C_1, C_2, \ldots, C_n \}$, where $C_i$ is the hypothesis ``the considered unit belongs to the class $i$''. In the classical belief function theory, the belief functions, also called the basic belief assignments, are defined by a mapping of the power set $2^\Theta$ onto $[0,1]$. The power set $2^\Theta$ is closed under the $\cup$ operator, and $\emptyset \in 2^\Theta$. In the extension proposed in the DSmT, generalized basic belief assignments are defined by a mapping of the hyper-power set $D^\Theta$ onto $[0,1]$, where the hyper-power set $D^\Theta$ is closed under both $\cup$ and $\cap$ operators. Consequently, we can manage finely the conflict of the experts and also take into account the tiles with more than one class. 

In the first section, we discuss and present different belief function models based on the power set and the hyper power set. These models try to answer our problem. We study these models also in the steps of combination and decision of the information fusion. These models allow, in a second section, to a general discussion on the difference between the DSmT and DST in terms of capacity to represent our problem and in terms of decision. Finally, we present an illustration of our proposed experts fusion on real sonar images, which represent a particularly uncertain environment.

\section{Our proposed Models}
In this section, we present five models taking into account the possible specificities of the application. First, we recall the principles of the DST and DSmT we apply here. Then we present a numerical example which illustrates the five proposed models presented afterward. The first three models are presented in the context of the DST, the fourth model in the context of the DSmT, and the fifth model in both contexts. 

\subsection*{\it Theory Bases}
\subsubsection*{\it Belief Function Models}
The belief functions or basic belief assignments $m$ are defined by the mapping of the power set $2^\Theta$ onto $[0,1]$, in the DST, and by the mapping of the hyper-power set $D^\Theta$ onto $[0,1]$, in the DSmT, with :

\begin{equation}
\label{close}
m(\emptyset)=0,
\end{equation}
and 
\begin{equation}
\label{normDST}
\sum_{X\in 2^\Theta} m(X)=1,
\end{equation}
in the DST, and 
\begin{equation}
\label{normDSmT}
\sum_{X\in D^\Theta} m(X)=1,
\end{equation}
in the DSmT, where $X$ is a given tile of the image. 

The equation (\ref{close}) allows that we assume a closed world \cite{Shafer76, Smarandache04}. We can define the belief function with only:
\begin{equation}
\label{open}
m(\emptyset)>0,
\end{equation}
and the world is open \cite{Smets90}. In a closed world, we can add one element in order to propose an open world.

These simple conditions in equation (\ref{close}) and (\ref{normDST}) or (\ref{close}) and (\ref{normDSmT}), give a large panel of definitions of the belief functions, which is one of the difficulties of the theory. The belief functions must therefore be chosen according to the intended application. 

In our case, the space of discernment $\Theta$ represents the different kind of sediments on sonar images, such as rock, sand, silt, cobble, ripple or shadow (that means no sediment information). The experts give their perception and belief according to their certainty. For instance, the expert can be moderately sure of his choice when he labels one part of the image as belonging to a certain class, and be totally doubtful on another part of the image. Moreover, on a considered tile, more than one sediment can be present. 

Consequently we have to take into account all these aspects of the applications. In order to simplify, we consider only two classes in the following: the rock referred as $A$, and the sand, referred as $B$. The proposed models can be easily extended, but their study is easier to understand with only two classes.

Hence, on certain tiles, $A$ and $B$ can be present for one or more experts. The belief functions have to take into account the certainty given by the experts (referred respectively as $c_A$ and $c_B$, two numbers in $[0,1]$) as well as the proportion of the kind of sediment in the tile $X$ (referred as $p_A$ and $p_B$, also two numbers in $[0,1]$). We have two interpretations of ``the expert believes $A$'': it can mean that the expert thinks that there is $A$ on $X$ and not $B$, or it can mean that the expert thinks that there is $A$ on $X$ and it can also have $B$ but he does not say anything about it. The first interpretation yields that hypotheses $A$ and $B$ are exclusive and with the second they are not exclusive. We only study the first case: $A$ and $B$ are exclusive. But on the tile $X$, the expert can also provide $A$ and $B$, in this case the two propositions ``the expert believes $A$'' and ``the expert believes $A$ and $B$'' are not exclusive.

\subsubsection*{\it Combination rules}
Many combination rules have been proposed these last years in the context of the belief function theory (\cite{Yager87, Dubois88, Smets90, Smets93, Smarandache04, Smarandache05}, {\it etc.}). In the context of the DST, the combination rule most used today seems to be the conjunctive consensus rule given by \cite{Smets90} for all $X \in 2^\Theta$ by:
\begin{eqnarray}
\label{consensus}
m(X)=\sum_{Y_1 \cap ... \cap Y_M = X} \prod_{j=1}^M m_j(Y_j),
\end{eqnarray}
where $Y_j \in 2^\Theta$ is the response of the expert $j$, and $m_j(Y_j)$ the associated belief function.

In the context of the DSmT, the conjunctive consensus rule can be used for all $X \in D^\Theta$ and $Y \in D^\Theta$. If we want to take the decision only on the elements in $\Theta$, some rules propose to redistribute the conflict on these elements. The most accomplished rule to provide that is the PCR5 given in \cite{Smarandache05} for two experts and for $X\in D^\Theta$, $X\neq \emptyset$ by:
\begin{eqnarray}
\label{DSmTcombination}
\begin{array}{l}
m_{PCR5}(X)=m_{12}(X)+\\
\displaystyle
\sum_{\begin{array}{l}
\scriptstyle Y\in D^\Theta, \\
\scriptstyle c(X\cap Y)=\emptyset
\end{array}} \left(\frac{m_1(X)^2 m_2(Y)}{m_1(X)+m_2(Y)}+\frac{m_2(X)^2 m_1(Y)}{m_2(X)+m_1(Y)}\right),
\end{array}
\end{eqnarray}
where $m_{12}(.)$ is the conjunctive consensus rule given by the equation (\ref{consensus}), \linebreak $c(X\cap Y)$ is the conjunctive normal form of $X\cap Y$ and the denominators are not null. We can easily generalize this rule for $M$ experts, for $X\in D^\Theta$, $X\neq \emptyset$ :
\begin{eqnarray}
\label{GeneDSmTcombination}
\displaystyle m_{PCR6}(X) &= & \displaystyle m(X) + \\
  \nonumber & & \!\!\!\!\!\!\!\!\!\!\sum_{i=1}^M m_i(X)^2
  \!\!\!\!\!\!\!\!\!\!\!\!\!\!\!\!\!\!\!\! \sum_{\begin{array}{c}
      \scriptstyle {\displaystyle \mathop{\cap}_{k=1}^{M\!-\!1}} Y_{\sigma_i(k)} \cap X \equiv \emptyset \\
      \scriptstyle (Y_{\sigma_i(1)},...,Y_{\sigma_i(M\!-\!1)})\in (D^\Theta)^{M\!-\!1}
  \end{array}}
  \!\!\!\!\!
  \left(\!\!\frac{\displaystyle \prod_{j=1}^{M\!-\!1} m_{\sigma_i(j)}(Y_{\sigma_i(j)})}
       {\displaystyle m_i(X) \!+\! \sum_{j=1}^{M\!-\!1}
  m_{\sigma_i(j)}(Y_{\sigma_i(j)})}\!\!\right)\!\!,
\end{eqnarray}
where $\sigma_i$ counts from 1 to $M$ avoiding $i$:
\begin{eqnarray}
\label{sigma}
\left\{
\begin{array}{ll}
\sigma_i(j)=j &\mbox{if~} j<i,\\
\sigma_i(j)=j+1 &\mbox{if~} j\geq i,\\
\end{array}
\right.
\end{eqnarray}
$m_i(X)+\displaystyle \sum_{j=1}^{M-1} m_{\sigma_i(j)}(Y_{\sigma_i(j)}) \neq 0$, and $m$ is the conjunctive consensus rule given by the equation (\ref{consensus}). 

The comparison of all the combination rules is not the purpose of this paper. Consequently, we use here the equation (\ref{consensus}) in the context of the DST and the equation (\ref{GeneDSmTcombination}) in the context of the DSmT.

\subsubsection*{\it Decision rules}

The decision is a difficult task. No measures are able to provide the best decision in all the cases. Generally, we consider the maximum of one of the three functions: credibility, plausibility, and pignistic probability. 

In the context of the DST, the credibility function is given for all $X \in 2^\Theta$ by:
\begin{eqnarray}
\bel(X)=\sum_{Y \in 2^X, Y \neq \emptyset} m(Y).
\end{eqnarray}
The plausibility function is given for all $X \in 2^\Theta$ by:
\begin{eqnarray}
\pl(X)=\sum_{Y \in 2^\Theta, Y\cap X \neq \emptyset} m(Y)=bel(\Theta)-bel(X^c),
\end{eqnarray}
where $X^c$ is the complementary of $X$. The pignistic probability, introduced by \cite{Smets90b}, is here given for all $X \in 2^\Theta$, with $X \neq \emptyset$ by:
\begin{eqnarray}
\betP(X)=\sum_{Y \in 2^\Theta, Y \neq \emptyset} \frac{|X \cap Y|}{|Y|} \frac{m(Y)}{1-m(\emptyset)}.
\end{eqnarray}
Generally the maximum of these functions is taken on the elements in $\Theta$, but we will give the values on all the focal elements.

In the context of the DSmT the corresponding generalized functions have been proposed \cite{Dezert04, Smarandache04}.
The generalized credibility $Bel$ is defined by:
\begin{eqnarray}
\Bel(X)=\sum_{Y \in D^X} m(Y)
\end{eqnarray}
The generalized plausibility $Pl$ is defined by:
\begin{eqnarray}
\Pl(X)=\sum_{Y \in D^\Theta, X \cap Y\neq \emptyset} m(Y)
\end{eqnarray}
The generalized pignistic probability is given for all $X \in D^\Theta$, with $X \neq \emptyset$ is defined by:
\begin{eqnarray}
\GPT(X)=\sum_{Y \in D^\Theta, Y \neq \emptyset} \frac{{\cal C_M}(X \cap Y)}{{\cal C_M}(Y)} m(Y),
\end{eqnarray}
where ${\cal C_M}(X)$ is the DSm cardinality corresponding to the number of parts of $X$ in the Venn diagram of the problem \cite{Dezert04, Smarandache04}.

If the credibility function provides a pessimist decision, the plausibility function is often too optimist. The pignistic probability is often taken as a compromise. We present the three functions for our models.

\subsection*{\it Numerical and illustrative example}
Consider two experts providing their opinion on the tile $X$. The first expert says that on tile $X$ there is some rock $A$ with a certainty equal to 0.6. Hence for this first expert we have : $p_A=1$, $p_B=0$, and $c_A=0.6$. The second expert thinks that there are 50\% of rock and 50\% of sand on the considered tile $X$ with a respective certainty of 0.6 and 0.4. Hence for the second expert we have: $p_A=0.5$, $p_B=0.5$, $c_A=0.6$ and $c_B=0.4$. We illustrate all our proposed models with this numerical exemple.

\subsection*{\it Model $M_1$}
If we consider the space of discernment given by $\Theta=\{A,B\}$, we can define a belief function by:
\begin{eqnarray}
\begin{array}{l}
\mbox{if the expert says $A$:}\\
  \left\{
  \begin{array}{l}
  m(A)=c_A, \\
  m(A \cup B)=1-c_A,
  \end{array}
  \right. \\
  \\
\mbox{if the expert says $B$:}\\
	\left\{
  \begin{array}{l}
  m(B)=c_B, \\
  m(A \cup B)=1-c_B.
  \end{array}
  \right.
\end{array}
\end{eqnarray}
In this case, it is natural to distribute $1-c_A$ and $1-c_B$ on $A\cup B$ which represent the ignorance. 

This model takes into account the certainty given by the expert but the space of discernment does not consider the possible heterogeneity of the given tile $X$. Consequently, we have to add another focal element meaning that there are two classes $A$ and $B$ on $X$. In the context of the Dempster-Shafer theory, we can call this focal element $C$ and the space of discernment is given by $\Theta=\{A,B,C\}$, and the power set is given by $2^\Theta=\{\emptyset,A,B,A\cup B,C, A\cup C, B \cup C, A\cup B \cup C\}$. Hence we can define our first model $M_1$ for our application by:
\begin{eqnarray}
\label{M1}
\begin{array}{l}
\mbox{if the expert says $A$:}\\
  \left\{
  \begin{array}{l}
  m(A)=c_A, \\
  m(A \cup B \cup C)=1-c_A,
  \end{array}
  \right. \\
  \\
\mbox{if the expert says $B$:}\\
	\left\{
  \begin{array}{l}
  m(B)=c_B, \\
  m(A \cup B \cup C)=1-c_B,
  \end{array}
  \right.  \\
  \\
\mbox{if the expert says $C$:}\\
  \left\{
  \begin{array}{l}
  m(C)=p_A . c_A + p_B . c_B, \\
  m(A \cup B \cup C)=1-(p_A . c_A + p_B . c_B).
  \end{array}
  \right.
\end{array}
\end{eqnarray}

On our numerical example, we obtain:
\begin{eqnarray*}
  \begin{array}{|c|c|c|c|c|}
  \hline
   & A & B & C & A \cup B \cup C \\
  \hline
   m_1 & 0.6 & 0 & 0 & 0.4 \\
  \hline
   m_2 & 0 & 0 & 0.5 & 0.5 \\
  \hline  
  \end{array}
\end{eqnarray*}
Hence for the consensus combination for the model $M_1$, the belief function $m_{12}$, the credibility, the plausibility and the pignistic probability are given by:
\begin{eqnarray*}
  \begin{array}{|c|c|c|c|c|}
  \hline
  element & m_{12} & \bel & \pl & \betP \\
  \hline
\emptyset & 0.3&  0  & 0 & - \\
  \hline
A & 0.3 & 0.3 & 0.5 & 0.5238 \\
  \hline
B & 0 & 0 & 0.2 & 0.0952 \\
  \hline
A\cup B & 0 &0.3 & 0.5 & 0.6190 \\
  \hline
C & 0.2 & 0.2 & 0.4 & 0.3810\\
  \hline
A \cup C & 0 & 0.5 & 0.7 & 0.9048 \\
  \hline
B\cup C & 0 &0.2  & 0.4 & 0.4762 \\
  \hline
A\cup B \cup C & 0.2 & 0.7 & 0.7 & 1 \\
  \hline
  \end{array}
\end{eqnarray*}
Where:
\begin{eqnarray}
  m_{12}(\emptyset)=m_{12}(A \cap C)=0.30.
\end{eqnarray}
This belief function provides an ambiguity because the same mass is put on $A$, the rock, and $\emptyset$, the conflict. With the maximum of credibility, plausibility or pignistic probability this ambiguity is suppressed because these functions do not consider the empty set.

\subsection*{\it Model $M_2$}
In the first model $M_1$, the possible heterogeneity of the tile is taken into account. However, the ignorance is characterized by $A \cup B \cup C$ and not by $A \cup B$ anymore, and class $C$ represents the situation when the two classes $A$ and $B$ are on $X$. Consequently $A \cup B \cup C$ could be equal to $A \cup B$, and we can propose another model $M_2$ given by:
\begin{eqnarray}
\label{M2}
\begin{array}{l}
\mbox{if the expert says $A$:}\\
  \left\{
  \begin{array}{l}
  m(A)=c_A, \\
  m(A \cup B)=1-c_A,
  \end{array}
  \right. \\
  \\
\mbox{if the expert says $B$:}\\
	\left\{
  \begin{array}{l}
  m(B)=c_B, \\
  m(A \cup B)=1-c_B,
  \end{array}
  \right.  \\
  \\
\mbox{if the expert says $C$:}\\
  \left\{
  \begin{array}{l}
  m(C)=p_A . c_A + p_B . c_B, \\
  m(A \cup B)=1-(p_A . c_A + p_B . c_B).
  \end{array}
  \right.
\end{array}
\end{eqnarray}

On our numerical example, we have:

\begin{eqnarray*}
  \begin{array}{|c|c|c|c|c|}
  \hline
   & A & B & C & A \cup B\\
  \hline
   m_1 & 0.6 & 0 & 0 & 0.4 \\
  \hline
   m_2 & 0 & 0 & 0.5 & 0.5 \\
  \hline  
  \end{array}
\end{eqnarray*}

In this model $M_2$ the ignorance is partial and the conjunctive consensus rule, the credibility, the plausibility and the pignistic probability are given by:
\begin{eqnarray*}
  \begin{array}{|c|c|c|c|c|}
  \hline
  element & m_{12}& \bel & \pl & \betP \\
  \hline
\emptyset & 0.5 & 0  & 0 & - \\
  \hline
A & 0.3 & 0.3 & 0.3 & 0.6 \\
  \hline
B & 0.2 &0.2 & 0.2 & 0.4 \\
  \hline
A\cup B &0 & 0.5 & 0.5 & 1 \\
  \hline
C & 0 & 0 & 0& 0\\
  \hline
A \cup C & 0 & 0.3 & 0.3 & 0.6 \\
  \hline
B\cup C & 0 & 0.2  & 0.2 & 0.4 \\
  \hline
A\cup B \cup C & 0 & 0.5 & 0.5 & 1 \\
  \hline
  \end{array}
\end{eqnarray*}
where 
\begin{eqnarray}
  m_{12}(\emptyset)=m_{12}(A \cap C)+m_{12}(C \cap (A\cup B))=0.30+0.2=0.5.
\end{eqnarray}

The previous ambiguity in $M_1$ between $A$ (the rock) and $\emptyset$ (the conflict) is still present with a belief on $\emptyset$ higher than $A$. Moreover, in this model the mass on $C$ is null!

These models $M_1$ and $M_2$ are different because in the DST the classes $A$, $B$ and $C$ are supposed to be exclusive. Indeed, the fact that the power set $2^\Theta$ is not closed under $\cap$ operator leads to the exclusivity of the classes.

\subsection*{\it Model $M_3$}
\begin{figure}[htb]
\begin{center}
\includegraphics[height=5cm]{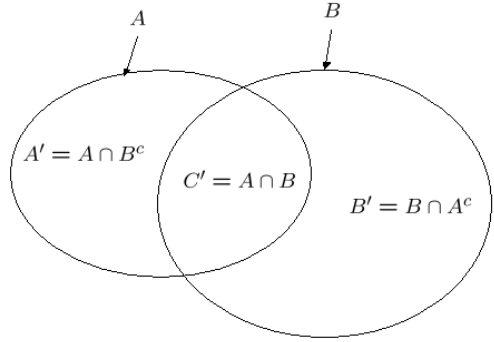}
\end{center}
\vspace{-0.5cm}
\caption{Notation of the intersection of two classes $A$ and $B$.}
\label{AetB}
\end{figure}

In our application, $A$, $B$ and $C$ cannot be considered exclusive on $X$. In order to propose a model following the DST, we have to study exclusive classes only. Hence, in our application, we can consider a space of discernment of three exclusive classes $\Theta=\{A \cap B^c, B\cap A^c, A \cap B\}=\{A', B', C'\}$, following the notations given on the figure \ref{AetB}.

Hence, we can propose a new model $M_3$ given by:
\begin{eqnarray}
\label{M3}
\begin{array}{l}
\mbox{if the expert says $A$:} \\
  \left\{
  \begin{array}{l}
  m(A'\cup C')=c_A, \\
  m(A' \cup B' \cup C')=1-c_A,
  \end{array}
  \right. \\
  \\
\mbox{if the expert says $B$:}\\
	\left\{
  \begin{array}{l}
  m(B'\cup C')=c_B, \\
  m(A' \cup B' \cup C')=1-c_B,
  \end{array}
  \right.  \\
  \\
\mbox{if the expert says $C$:}\\
  \left\{
  \begin{array}{l}
  m(C')=p_A . c_A + p_B . c_B, \\
  m(A' \cup B' \cup C')=1-(p_A . c_A + p_B . c_B).
  \end{array}
  \right.
\end{array}
\end{eqnarray}
Note that $A' \cup B' \cup C'=A\cup B$. On our numerical example we obtain:
\begin{eqnarray*}
  \begin{array}{|c|c|c|c|c|}
  \hline
   & A'\cup C' & B' \cup C' & C' & A' \cup B' \cup C'\\
  \hline
   m_1 & 0.6 & 0 & 0 & 0.4 \\
  \hline
   m_2 & 0 & 0 & 0.5 & 0.5 \\
  \hline  
  \end{array}
\end{eqnarray*}

Hence, the conjunctive consensus rule, the credibility, the plausibility and the pignistic probability are given by:
\begin{eqnarray*}
  \begin{array}{|c|c|c|c|c|}
  \hline
  element & m_{12} & \bel & \pl & \betP \\
  \hline
\emptyset & 0 & 0 & 0 & - \\
  \hline
A'=A \cap B^c &0 & 0 & 0.5 & 0.2167 \\
  \hline
B'=B\cap A^c & 0 & 0& 0.2 & 0.0667 \\
  \hline
A'\cup B'=(A \cap B^c)\cup (B\cap A^c)& 0& 0 & 0.5 & 0.2833 \\
  \hline
C'=A \cap B & 0.5 & 0.5 & 1 & 0.7167\\
  \hline
A' \cup C'=A & 0.3 & 0.8 & 1 & 0.9333 \\
  \hline
B'\cup C'=B & 0 & 0.5  & 1 & 0.7833 \\
  \hline
A'\cup B' \cup C'=A\cup B & 0.2 & 1 & 1 & 1 \\
  \hline
  \end{array}
\end{eqnarray*}
where
\begin{eqnarray}
m_{12}(C')=m_{12}(A\cap B)=0.2+0.3=0.5.
\end{eqnarray}

On this example, with this model $M_3$ the decision will be $A$ with the maximum of pignistic probability. But the decision could {\it a priori} be taken also on $C'=A\cap B$ because $m_{12}(C')$ is the highest. We show however in the discussion section that it is not possible.

\subsection*{\it Model $M_4$}
In the context of the DSmT, we can write $C=A\cap B$ and easily propose a fourth model $M_4$, without any consideration on the exclusivity of the classes, given by:

\begin{eqnarray}
\label{M4}
\begin{array}{l}
\mbox{if the expert says $A$:}\\
  \left\{
  \begin{array}{l}
  m(A)=c_A, \\
  m(A \cup B)=1-c_A,
  \end{array}
  \right. \\
  \\
\mbox{if the expert says $B$:}\\
	\left\{
  \begin{array}{l}
  m(B)=c_B, \\
  m(A \cup B)=1-c_B,
  \end{array}
  \right.  \\
  \\
\mbox{if the expert says $A\cap B$:}\\
  \left\{
  \begin{array}{l}
  m(A \cap B)=p_A . c_A + p_B . c_B, \\
  m(A \cup B)=1-(p_A . c_A + p_B . c_B).
  \end{array}
  \right.
\end{array}
\end{eqnarray}
This last model $M_4$ allows to represent our problem without adding an artificial class $C$. Thus, the model $M_4$ based on the DSmT gives:

\begin{eqnarray*}
  \begin{array}{|c|c|c|c|c|}
  \hline
   & A & B & A \cap B & A \cup B \\
  \hline
   m_1 & 0.6 & 0 & 0 & 0.4 \\
  \hline
   m_2 & 0 & 0 & 0.5 & 0.5 \\
  \hline  
  \end{array}
\end{eqnarray*}

The obtained mass $m_{12}$ with the conjunctive consensus yields:
  
\begin{eqnarray}
\label{M4consensus}
  \begin{array}{l}
  m_{12}(A)=0.30, \\
  m_{12}(B)=0, \\
  m_{12}(A \cap B)=m_1(A)m_2(A\cap B)+  m_1(A\cup B)m_2(A\cap B)\\
  \quad \quad \quad \quad \quad=0.30+0.20=0.5,\\
  m_{12}(A\cup B)=0.20. \\
  \end{array}
\end{eqnarray}

These results are exactly the same for the model $M_3$. These two models do not present ambiguity and show that the mass on $A\cap B$ (rock and sand) is the highest. 

The  generalized credibility, the  generalized plausibility and the generalized pignistic probability are given by:
\begin{eqnarray*}
  \begin{array}{|c|c|c|c|c|}
  \hline
  element  & m_{12} & \Bel & \Pl & \GPT \\
  \hline
\emptyset & 0 & 0  & 0 & - \\
  \hline
A & 0.3&  0.8 & 1 & 0.9333 \\
  \hline
B & 0 & 0.5 & 0.7& 0.7833 \\
  \hline
A\cap B& 0.5& 0.5 & 1 & 0.7167 \\
  \hline
A\cup B & 0.2 & 1  & 1 & 1 \\
  \hline
  \end{array}
\end{eqnarray*}

Like the model $M_3$, on this example, the decision will be $A$ with the maximum of pignistic probability criteria. But here also the maximum of $m_{12}$ is reached for $A\cap B=C'$.

If we want to consider only the kind of possible sediments $A$ and $B$ and not also the conjunctions, we can use a proportional conflict redistribution rules such as the PCR5 proposed in \cite{Smarandache05}. Consequently we have $x=0.3.(0.5/0.3)=0.5$ and $y=0$, and the PCR5 rule provides:
\begin{eqnarray}
\label{M4PCR5}
  \begin{array}{l}
  m_{PCR5}(A)=0.30+0.5=0.8, \\
  m_{PCR5}(B)=0, \\
  m_{PCR5}(A\cup B)=0.20. \\
  \end{array}
\end{eqnarray}

The credibility, the plausibility and the pignistic probability are given by:
\begin{eqnarray*}
  \begin{array}{|c|c|c|c|c|}
  \hline
  element  & m_{PCR5} & \bel & \pl & \betP \\
  \hline
\emptyset & 0 & 0  & 0 & - \\
  \hline
A & 0.8&  0.8 & 1 & 0.9 \\
  \hline
B & 0 & 0& 0.2& 0.1 \\
  \hline
A\cup B & 0.2 & 1  & 1 & 1 \\
  \hline
  \end{array}
\end{eqnarray*}
On this numerical example, the decision will be the same than the consensus rule, here the maximum of pignistic probability is reached for $A$ (rock). In the next section we see that is not always the case.

\subsection*{\it Model $M_5$}
Another model $M_5$ which can be used in both the DST and the DSmT is given considering only one belief function according to the proportion by:
\begin{eqnarray}
\label{M5}
\begin{array}{l}
  \left\{
  \begin{array}{l}
  m(A)=p_A.c_A, \\
  m(B)=p_B.c_B,\\
  m(A \cup B)=1-(p_A . c_A + p_B . c_B).
  \end{array}
  \right. \\
\end{array}
\end{eqnarray}
If for one expert, the tile contains only $A$, $p_A=1$, and $m(B)=0$. If for another expert, the tile contains $A$ and $B$, we take into account the certainty and proportion of the two sediments but not only on one focal element. Consequently, we have simply:
\begin{eqnarray*}
  \begin{array}{|c|c|c|c|}
  \hline
   & A & B & A \cup B \\
  \hline
   m_1 & 0.6 & 0  & 0.4 \\
  \hline
   m_2 & 0.3& 0.2 & 0.5 \\
  \hline  
  \end{array}
\end{eqnarray*}

In the DST context, the consensus rule, the credibility, the plausibility and the pignistic probability are given by:
\begin{eqnarray*}
  \begin{array}{|c|c|c|c|c|}
  \hline
  element  & m_{12} & \bel & \pl & \betP \\
  \hline
\emptyset & 0.12 & 0  & 0 & - \\
  \hline
A & 0.6&  0.6 & 0.8 & 0.7955 \\
  \hline
B & 0.08 & 0.08& 0.28& 0.2045 \\
  \hline
A\cup B & 0.2 & 0.88  & 0.88 & 1 \\
  \hline
  \end{array}
\end{eqnarray*}
In this case we do not have the plausibility to decide on $A\cap B$, because the conflict is on $\emptyset$.

In the DSmT context, the consensus rule, the generalized credibility, the generalized plausibility and the generalized pignistic probability are given by:
\begin{eqnarray*}
  \begin{array}{|c|c|c|c|c|}
  \hline
  element  & m_{12} & \Bel & \Pl & \GPT \\
  \hline
\emptyset & 0 & 0  & 0 & - \\
  \hline
A & 0.6&  0.72 & 0.92 & 0.8933 \\
  \hline
B & 0.08 & 0.2& 0.4& 0.6333 \\
  \hline
A\cap B& 0.12& 0.12 & 1 & 0.5267 \\
  \hline
A\cup B & 0.2 & 1  & 1 & 1 \\
  \hline
  \end{array}
\end{eqnarray*}
The decision with the maximum of pignistic probability criteria is still $A$.

The PCR5 rule provides:
\begin{eqnarray*}
  \begin{array}{|c|c|c|c|c|}
  \hline
  element  & m_{PCR5} & \bel & \pl & \betP \\
  \hline
\emptyset & 0 & 0  & 0 & - \\
  \hline
A & 0.69&  0.69 & 0.89 & 0.79 \\
  \hline
B & 0.11 & 0.11& 0.31& 0.21 \\
  \hline
A\cup B & 0.2 & 1  & 1 & 1 \\
  \hline
  \end{array}
\end{eqnarray*}
where
\begin{eqnarray*}
  \begin{array}{l}
  m_{PCR5}(A)=0.60+0.09=0.69, \\
  m_{PCR5}(B)=0.08+0.03=0.11. \\
  \end{array}
\end{eqnarray*}
With this model and example the PCR5 rule, the decision will be also $A$, and we do not have difference between the consensus rules in the DST and DSmT.

\section{Discussion}

We have build, in the previous section, the models $M_1$, $M_2$, $M_3$, $M_4$, and $M_5$ in the DSmT case in order to take into account the decision considering also $A\cap B$ (``there is rock and sand on the tile''). In fact only the $M_1$ and $M_2$ models can do it. Model $M_2$ can do it only if both experts say $A\cap B$. These two models assume that $A$, $B$ and $A\cap B$ are exclusive. Of course this assumption is false. For the models $M_3$, $M_4$ and $M_5$, we have to take the decision on the credibilities, plausibilities or pignistic probabilities, but these three functions for $A\cap B$ cannot be higher than $A$ or $B$ (or for $C'$ than $A'\cup C'$ and $B'\cup C'$ with the notations of the model $M_3$). Indeed for all $x \in A\cap B$, $x\in A$ and $x \in B$, so for all $X \subseteq Y$:
\begin{eqnarray*}
  \begin{array}{l}
\bel(X) \leq \bel(Y),\\
\pl(X) \leq \pl(Y),\\
\betP(X) \leq \betP(Y),\\
\Bel(X) \leq \Bel(Y),\\
\Pl(X) \leq \Pl(Y),\\
\GPT(X) \leq \GPT(Y).\\
  \end{array}
\end{eqnarray*}

Hence, our first problem is not solved: we can never choose $A\cap B$ with the maximum of credibility, plausibility or pignistic probability. If the two experts think that the considered tile contains rock and sand ($A \cap B$), then the pignistic probabilities are equal. However the belief on $A\cap B$ can be the highest (see the example on the models $M_3$ and $M_4$). The limits of the decision rules are reached in this case.

We have seen that we can describe our problem both in the DST and the DSmT context. The DSmT is more adapted to modelize the belief on $A\cap B$ for example with the model $M_4$, but model $M_3$ with the DST can provide exactly the same belief on $A$, $B$ and $A\cap B$. Consequently, the only difference we can expect on the decision comes from the combination rules. In the presented numerical example, the decisions are the same: we choose $A$. 

\subsection*{\it An example of decision instability}

Take another example with this last model $M_5$: The first expert
provides: $p_A=0.5$, $p_B=0.5$, $c_A=0.6$ and $c_B=0.4$, and the
second expert provides: $p_A=0.5$, $p_B=0.5$, $c_A=0.86$ and
$c_B=1$. We want take a decision only on $A$ or $B$. Hence we have:
\begin{eqnarray*}
  \begin{array}{|c|c|c|c|}
  \hline
   & A & B & A \cup B \\
  \hline
   m_1 & 0.3 & 0.2  & 0.5 \\
  \hline
   m_2 & 0.43& 0.5 & 0.07 \\
  \hline  
  \end{array}
\end{eqnarray*}

For $M_5$ on the DST context:
\begin{eqnarray*}
  \begin{array}{|c|c|c|c|c|}
  \hline
  element  & m_{12} & \bel & \pl & \betP \\
  \hline
\emptyset & 0.236 & 0  & 0 & - \\
  \hline
A & 0.365&  0.365 & 0.4 & 0.5007 \\
  \hline
B & 0.364 & 0.364 & 0.399& 0.4993 \\
  \hline
A\cup B & 0.035 & 0.764  & 0.764 & 1 \\
  \hline
  \end{array}
\end{eqnarray*}

$M_5$ with PCR5 gives (with the partial conflicts: $x_1=0.0562$, $y_1=0.0937$, $x_2=0.0587$ and $y_2=0.0937$):
\begin{eqnarray*}
  \begin{array}{|c|c|c|c|c|}
  \hline
  element  & m_{PCR5} & \bel & \pl & \betP \\
  \hline
\emptyset & 0 & 0  & 0 & - \\
  \hline
A & 0.479948&  0.479 & 0.5149 & 0.4974 \\
  \hline
B & 0.485052 & 0.485& 0.5202& 0.5026 \\
  \hline
A\cup B & 0.035 & 1  & 1 & 1 \\
  \hline
  \end{array}
\end{eqnarray*}
This last example shows that we have a difference between the DST and
the DSmT, but what is the best solution? With the DST we choose $A$
and with the DSmT we choose $B$. We can show that the decision will be
the same in the most of the case (about 99.4\%).

\subsection*{\it Stability of decision process}

The space where experts can define their opinions on which $n$ classes are present in a given tile is a part of $[0,1]^n$: $\mathcal{E} = [0,1]^n\cap (\displaystyle \sum_{X\in\Theta} m(X)\leq 1)$. In order to study the different combination rules, and the situations where they differ, we use a Monte Carlo method, considering the weights $p_A$, $c_A$, $p_B$, $c_B$, \ldots, as uniform variables, filtering them by the condition $\displaystyle \sum_{X\in\Theta}p_Xc_X\leq 1$ for one expert.

Thus, we measure the proportion of situations where decision differs between the consensus combination rule, and the PCR5, where conflict is proportionally distributed.

We can not choose $A\cap B$, as the measure of $A\cap B$ is always
lower (or equal with probability 0) than the measure of $A$ or $B$. In
the case of two classes, $A\cup B$ is the ignorance, and is usually
excluded (as it always maximises $\bel$, $\pl$, $\betP$, $\Bel$, $\Pl$ and
$\GPT$). We restrict the possible choices to singletons, $A$, $B$,
etc. Therefore, it is equivalent to tag the tile by the most credible
class (maximal for $\bel$), the most plausible (maximal for $\pl$), the
most probable (maximal for $\betP$) or the heaviest (maximal for $m$),
as the only focal elements are singletons, $\Theta$ and $\emptyset$.

The only situation where the total order induced by the masses $m$ on
singletons can be modified is when the conflict is distributed on the
singletons, as is the case in the PCR5 method. \\

Thus, for two classes, the subspace where the decision is ``rock'' by
consensus rule is very similar to the subspace where the decision is
``rock'' by the PCR5 rule: only 0.6\% of the volume differ. For a higher
number of classes, the decision obtained by fusing the two
experts' opinions is much less stable:

\begin{center}
  \begin{tabular}{|l|c|c|c|c|c|c|}
    \hline
    number of classes & 2 & 3 & 4 & 5 & 6 & 7 \\
    \hline
    decision change & 0.6\% & 5.5\% & 9.1\% & 12.1\% & 14.6\% & 16.4\% \\
    \hline
  \end{tabular}
\end{center} 

Therefore, the specificity of PCR5 appears mostly with more than two
classes, and the different combination rules are nearly equivalent
when decision must be taken within two possible classes.

\begin{figure}[htb]
  \begin{center}
    \begin{tabular}{cc}
      \includegraphics[height=6cm]{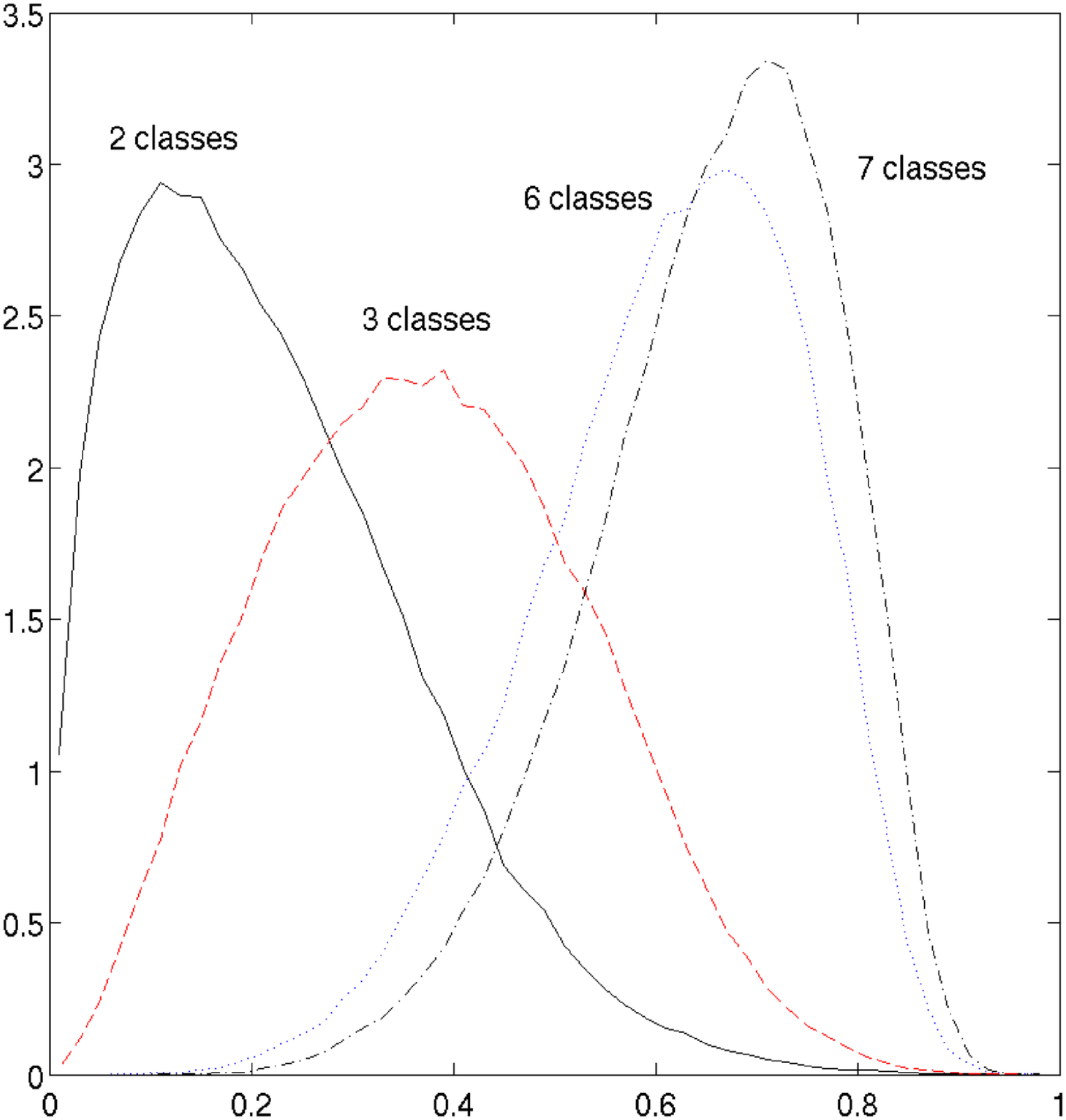} &
      \includegraphics[height=6cm]{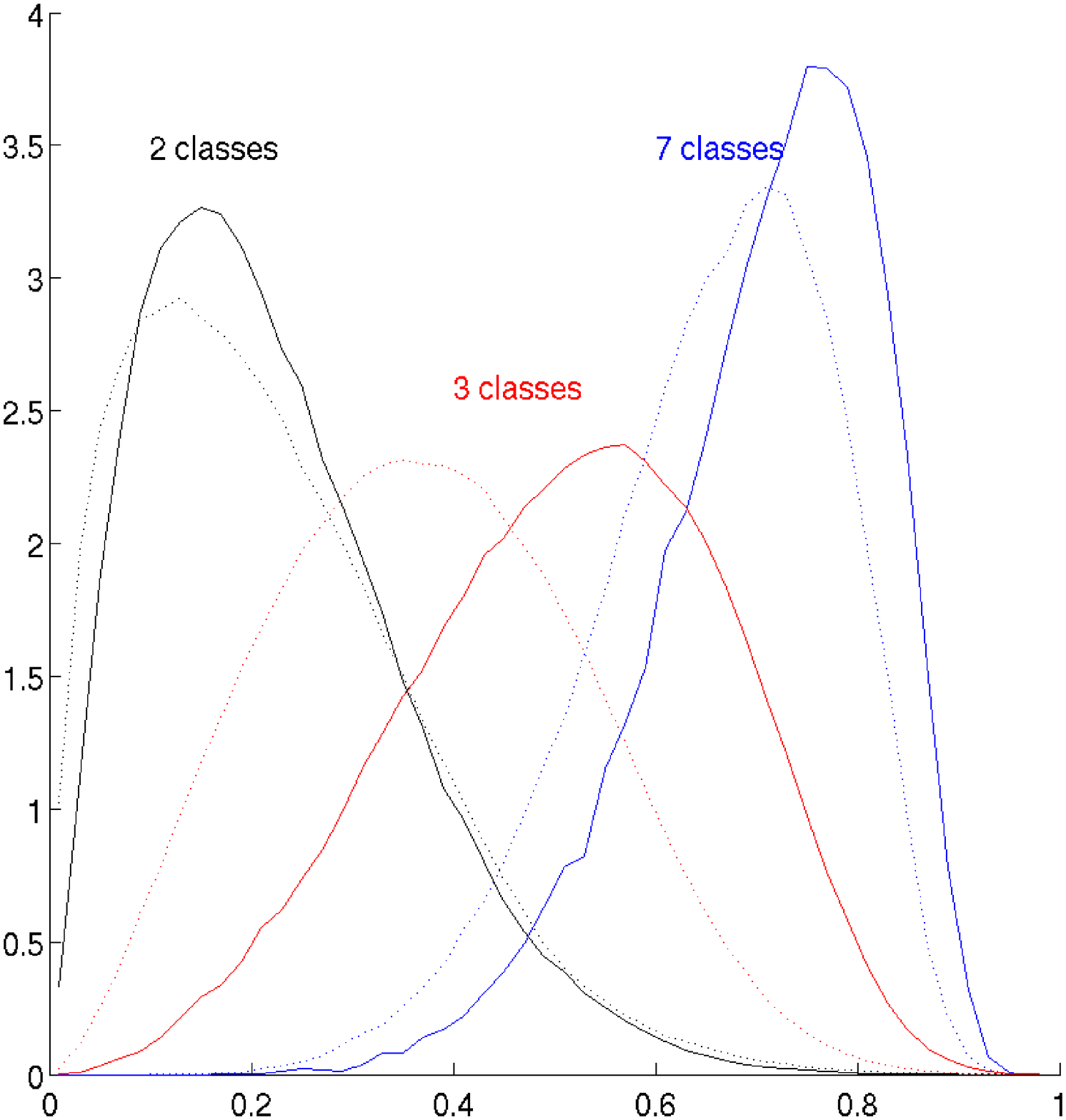} \\
    \end{tabular}
  \end{center}
  \vspace{-0.5cm}
  \caption{Density of conflict for (left) uniform random experts and (right)
    data with different decision between consensus and PCR5.}
  \label{conflictdensity}
\end{figure}

Left part of figure \ref{conflictdensity} shows the density of
conflict within $\mathcal{E}$, for a number of classes of 2, 3, 6 and
7. Right part shows how this distribution changes if we restrict
$\mathcal{E}$ to the cases where the decision changes between
consensus (dotted lines) and PCR5 (plain lines). Conflict is more
important in this subspace, mostly because a low conflict usually
means a clear decision: the measure on the best class is often very different
than measure on the second best class.

For the ``two experts and two classes'' case, it is difficult to characterize analytically the stability of the decision process. However, we can easily show that if $m_1(A)=m_2(B)$ or if $m_1(A)=m_1(B)$, the final decision does not depend on the chosen combination rule.

\section{Illustration}
\label{illustration}

\subsection*{\it Database}
Our database contains 40 sonar images provided by the GESMA (Groupe \linebreak d'Etudes Sous-Marines de l'Atlantique). These images were obtained with a Klein 5400 lateral sonar with a resolution of 20 to 30 cm in azimuth and 3 cm in range. The sea-bottom depth was between 15 m and 40 m.

\begin{figure}[htb]
\begin{center}
\includegraphics[height=5cm]{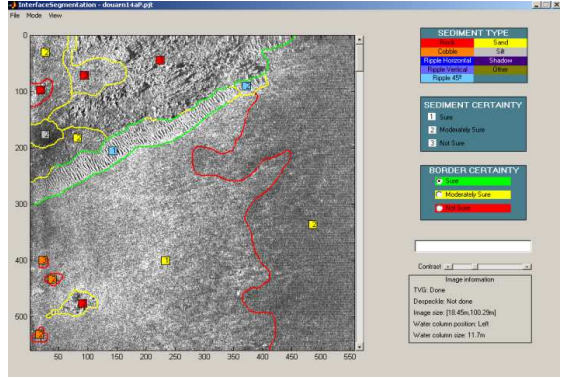}
\end{center}
\caption{Manual Segmentation Interface.}
\label{manual_seg}
\end{figure}

Two experts have manually segmented these images giving the kind of
sediment (rock, cobble, sand, silt, ripple (horizontal, vertical or at
45 degrees)), shadow or other (typically ships) parts on images,
helped by the manual segmentation interface presented in figure
\ref{manual_seg}. All sediments are given with a certainty level
(sure, moderately sure or not sure). Hence, every pixel of every image
is labeled as being either a certain type of sediment or a shadow or
other.

\subsection*{\it Results}
We note $A=$ rock, $B=$ cobble, $C=$ sand, $D=$ silt, $E=$ ripple,
$F=$ shadow and $G=$ other, hence we have seven classes and $\Theta=\{A,B,C,D,E,F,G\}$. We have applied the generalized model $M_5$ on tiles of size 32$\times$32 given by:
\begin{eqnarray}
\label{GeneralizedM5}
\begin{array}{l}
  \left\{
  \begin{array}{l}
  m(A)=p_{A1}.c_1+p_{A2}.c_2+p_{A3}.c_3, \, \mbox{for rock,}\\
  m(B)=p_{B1}.c_1+p_{B2}.c_2+p_{B3}.c_3, \, \mbox{for cobble,}\\
  m(C)=p_{C1}.c_1+p_{C2}.c_2+p_{C3}.c_3, \, \mbox{for ripple,}\\
  m(D)=p_{D1}.c_1+p_{D2}.c_2+p_{D3}.c_3, \, \mbox{for sand,}\\
  m(E)=p_{E1}.c_1+p_{E2}.c_2+p_{E3}.c_3, \, \mbox{for silt,}\\
  m(F)=p_{F1}.c_1+p_{F2}.c_2+p_{F3}.c_3, \, \mbox{for shadow,}\\
  m(G)=p_{G1}.c_1+p_{G2}.c_2+p_{G3}.c_3, \, \mbox{for other,}\\
  m(\Theta)=1-(m(A)+m(B)+m(C)+m(D)+m(E)+m(F)+m(G)),\\
  \end{array}
  \right. \\
\end{array}
\end{eqnarray}
where $c_1$, $c_2$ and $c_3$ are the weights associated to the
certitude respectively: ``sure'', ``moderately sure'' and ``not
sure''. The chosen weights are here: $c_1=2/3$, $c_2=1/2$ and
$c_3=1/3$. Indeed we have to consider the cases when the same kind of sediment (but with different certainties) is present on the same tile. The proportion of each sediment in the tile associated to these weights is noted, for instance for $A$: $p_{A1}$, $p_{A2}$ and $p_{A3}$. The
table \ref{Conflict} gives the conflict matrix of the two experts. We note that the most of conflict come from a difference of opinion between sand and silt. For instance, the expert 1 provides many tiles of sand when the expert 2 thinks that is silt (conflict induced of 0.0524). This conflict is explained by the difficulty for the experts to differentiate sand and silt that differ with only the intensity. Part of conflict comes also from the fact that ripples are hard to distinguish from sand or silt. Ripples, that is, sand or silt in a special configuration, is sometimes difficult to see on the images, and the ripples are most of the time visible in a global zone where sand or silt is present. Cobbles also yield conflicts, especially with sand, silt and rock: cobble is described by some small rocks on sand or silt. The total conflict between the two experts is 0.1209. Hence, our application does not present a large conflict.

\begin{table}
  \centering
  \begin{tabular}{cc}
    & Expert 2\\
    {\begin{tabular}{r}
	{\rotatebox[origin=lb]{90}{Expert 1}}\hspace*{-0.5cm}\\
    \end{tabular}} & 
    {\begin{tabular}{|c|c|c|c|c|c|c|c|}
	\hline
	& Rock & Cobble & Ripple & Sand & Silt  & Shadow & Other\\
	\hline
	Rock & - & 12.87 & 2.72 & 4.42 & 3.91 & 6.41 & 0.22\\
	\hline
	Cobble & 5.59 & - & 0.85 & 18.44 & 3.85 & 0.04 & 0\\
	\hline
	Ripple & 3.12 & 3.38 & - & 30.73 & 150.60 & 0.27 & 0.16 \\
	\hline
	Sand & 9.50 & 43.39 & 42.60 & - & 524.33 & 0.51 & 0.57\\
	\hline
	Silt & 6.42 & 27.05 & 36.22 & 258.98 & - & 2.60 & 0.11\\
	\hline
	Shadow & 3.82 & 0.15 & 2.13 & 1.38 & 0.50 & - & 0.41\\
	\hline
	Other & 0 & 0.20 & 0.10 & 0.35 & 0.31 & 0.14 &- \\
	\hline
    \end{tabular}}\\
  \end{tabular}
  \caption{Matrix of conflict ($\times 10^{4}$) between the two experts.}
\label{Conflict}
\end{table}

We have applied the consensus rule and the PCR5 rule with this model. The decision is given by the maximum of pignistic probability. In most of the cases the decisions taken by the two rules are the same. We note a difference only on 0.4657\% of the tiles. Indeed, we are in the seven classes case with only 0.1209 of conflict, the simulation given on the figure 3 show that we have few chance that the decisions differ.

\section{Conclusion}
In this paper we have proposed five different models in order to take into account two classical problems in uncertain image classification (for training or evaluation): the heterogeneity of the considered tiles and the certainty of the experts. These five models have been developed in the DST and DSmT contexts. The heterogeneity of the tile and the certainty of the expert can be easily taken into account in the models. However, if we want to have the plausibility of taking a decision on such a tile (with a conjunction $A\cap B$) the usual decision functions (credibility, plausibility and pignistic probability) are not sufficient: they cannot allow a such decision. We can take the decision on $A\cap B$ only if we consider the belief function and if the model provides a belief on $A\cap B$. 

We have also studied the decision according to the conflict and to the combination rules: conjunctive consensus rule and PCR5 rule. The decision (taken with the maximum of the credibility, the plausibility or the pignistic probability) is the same in most of the cases. For two experts, more classes leads to more conflict and to more cases giving a different decision with the different rules.

We have also illustrated one of the proposed models on real sonar images classified manually by two different experts. In this application the total conflict between the two experts is 0.1209 and we note a difference of decision only on 0.4657\% of the tiles.

We can easily generalize our models for three or more experts and use the generalized combination of the PCR5 given by the equation (\ref{GeneDSmTcombination}). Of course the conflict will be higher and the difference in the decision must be studied.

\vfill

\section*{}
{\bf ARNAUD MARTIN} is a teacher and researcher at the ENSIETA in the laboratory $E^3I^2$: EA3876, Brest, France. He received a PhD degree in Signal Processing (2001), and Master in Probability (1998). His research interests are mainly related to data fusion, data mining, signal processing especially for sonar and radar data. {\it E-mail}: Arnaud.Martin@ensieta.fr.

\section*{}
{\bf CHRISTOPHE OSSWALD} is a teacher and researcher at teh ENSIETA in the laboratory $E^3I^2$: EA3876, Brest, France. He received a PhD degree
in Mathematics and Computer Science (2003) and engineer graduated from Ecole Polytechnique (promotion 1994). His research interests are related to
classification, data fusion and (hyper)graphs theory. {\it E-mail}: Christophe.Osswald@ensieta.fr.

\end{document}